# Learning Stabilizing Control Policies for a Tensegrity Hopper with Augmented Random Search

Vladislav Kurenkov
*Center for Technologies in Robotics and Mechatronics Components*
*Innopolis University*
Innopolis, Russia
v.kurenkov@innopolis.ru

Hany Hamed
*Robotics Institute*
*Innopolis University*
Innopolis, Russia
h.hamed@innopolis.university

Sergei Savin
*Center for Technologies in Robotics and Mechatronics Components*
*Innopolis University*
Innopolis, Russia
s.savin@innopolis.ru

*Abstract*—In this paper, we consider tensegrity hopper - a novel tensegrity-based robot, capable of moving by hopping. The paper focuses on the design of the stabilizing control policies, which are obtained with Augmented Random Search method. In particular, we search for control policies which allow the hopper to maintain vertical stability after performing a single jump. It is demonstrated, that the hopper can maintain a vertical configuration, subject to the different initial conditions and with changing control frequency rates. In particular, lowering control frequency from 1000Hz in training to 500Hz in execution did not affect the success rate of the balancing task.

## I. Introduction

For a variety of applications, it is desirable to use soft, foldable and collision-resilient robots. For example, for space robotics the size and weight of the mechanism are crucial, and for landers the robustness is extremely important. In collaborative robotics an additional requirement is present: the minimization of risks of the robot dealing damage to the objects in the environment. All of those properties are well captured by tensegrity structures [1-4].

Tensegrity structures are often defined as consisting of struts, which experience compressive forces, cables which experience tensile forces, and bars which experience both [5]. Tensegrity structures have been used in design of planetary probes by NASA [2], where a structure with six links connected via elastic cables was used [6]. The structure was able to move by rolling, which was achieved by controlling the rest lengths of the cables (lengths of the cables, when no tensile forces are present). This is one of the main methods of achieving motions of tensegrity structures. Tensegrity structures have also been proposed as elements of walking robots and climbing, swimming mechanisms and deployable antennas [7-9].

This paper proposes a novel hopper tensegrity robot, consisting of two links. It is not a pure tensegrity structure, as one of the links experiences bending forces. The robot can move by using four actuated cables. Alternatively, all eight cables can be actuated. The control of such a robot is an open research question.

There are a number of open control problems in tensegrity robotics: trajectory design [7], inverse kinematics [10], feedback control [11], state estimation [12], and others. In the absence of well-established methods, one of the popular approaches is to use machine learning techniques for generating trajectories or control policies for the tensegrity structures. Current methods are mostly based on the use of Central Pattern Generators (CPG) [13-15]. These approaches, while limiting, allow controlling a variety of structures, including six bar rollers and spine-like robots [13-14].

In this paper, a more general approach, based on Augmented Random Search (ARS) is proposed. The rest of the paper is organized as follows. Next section provides a description of the state of the art in policy search methods. Section III provides description of the hopper robot. Section IV gives a description of the proposed policy search method based on ARS and section V shows results achieved with the proposed method. In particular, it shows that the robot can be stabilized in a vertical position, standing on a single node, after falling from a range of heights. It is also shown that the robot is robust to the changes in control update frequency.

## II. State of the Art in Policy Search methods

Recent rise of interest in reinforcement learning resulted in an extensive amount of new policy search methods for continuous control problems [17, 19, 20- 22, 26, 28]. Usually, reinforcement learning algorithms are divided into two major categories. First, value-based algorithms [24-25, 27, 31], estimate value function for a given policy, and then use this estimate to find a better policy, repeating this process until the optimal policy is found. Although, they are inherently unsuitable for continuous domains [30], several successful extensions were proposed to alleviate this limitation [20, 22, 30]. Another major branch is the policy gradient methods [16-17, 19, 21, 26, 28-29]. They directly search for the policy as opposed to the value-based methods, which naturally allows to handle continuous control problems.

Recently, simpler methods were shown to be competitive and achieve the same level of performance [18, 23]. One of

The research is supported by grant of the Russian Science Foundation (project No:19-79-10246).

such methods, Augmented Random Search [23], successfully finds linear policies that are competitive to the policies obtained by other deep reinforcement learning algorithms for continuous control problems. Even though these methods are less data-efficient, they are far easier to parallelize, that significantly reduces actual training time. Moreover, they are believed to solve long-horizon problems and sparse reward settings more effectively [18].

## III. ROBOT DESCRIPTION

In this paper, we consider a structure based on two rigid bodies connected by a system of 8 cables, as shown in Figure 1.

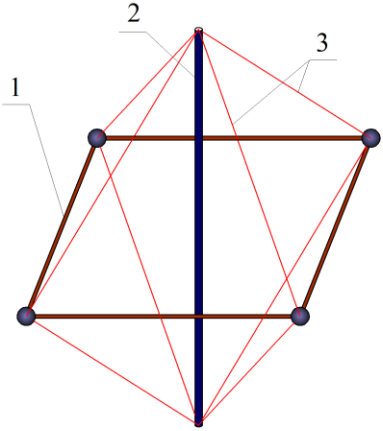

Fig. 1. Eight cable tensegrity jumber; 1 – frame link, 2 – leg link, 3 – cables.

The hopper can be built in such a way that the leg link (see Figure 1) would have much smaller mass than the frame link, allowing it to quickly move from one position to the next during the flight phase of the robot motion, without significant influence (due to dynamic coupling via elastic forces) on the trajectory of the frame link. This makes it similar to the well-known spring-loaded inverted pendulum (SLIP) model.

## IV. POLICY SEARCH

### A. The Policy Search Problem

The goal of the controller is to make the tensegrity structure stand in the vertical position (which would be an unstable equilibrium in the absence of the control actions) after falling from a given height; when standing, the structure contacts the ground via only one of its nodes. For training simulations, the structure was placed 1m above the ground.

Formally, we frame the problem as a finite-horizon Markov Decision Process: <$S$, $A$, $P$, $R$>, where $S$ is the state space, $A$ is action space, $P$ is transition function, and $R$ is a reward function. The state space $S$ is represented as a set of tuples. Every tuple consists of cable lengths, current position and current velocity of every node. This representation results in an observation space of 44 dimensions. The action space $A$ is represented as a set of tuples, where every tuple describes a desired change in the target length of every cable, resulting in the eight dimensions continuous control problem. The transition function $P$ is unknown, which leads to a model-free reinforcement learning setting.

The reward function $R$ is defined as to achieve the stable standing on one of the nodes after falling down from the specified height. Every time step, the agent is given a positive reward if all of the following criteria holds true. Firstly, the angle between the leg link and the ground should stay in the interval of [-20, 20] degrees. Secondly, the angle between the frame link and the ground should stay in the interval of [-40, 40] degrees. If these conditions do not hold, we terminate the episode. Therefore, the more time steps the agent suffices these criteria, the more reward it gets. We limit the horizon by 20000 time steps. Every step occurs exactly at 1000Hz rate. Thus every episode is limited by 20 seconds.

We wish to find policy parameters $\theta^*$ that maximize the expected return $E[J(\theta)]$:

$$\theta^* = \arg\max(E[J(\theta)]) \qquad (1)$$

$$J(\theta) = \sum_{t=1}^{N} R(s_t, u_t), \quad u_t \sim \pi(u|s_{t-1}, \theta) \qquad (2)$$

where $u_t$ is randomly sampled out of the policy $\pi(u|s_{t-1}, \theta)$.

### B. Augmented Random Search

To solve the proposed optimization problem, we use Augmented Random Search [32]. This method has shown to be competitive to the state-of-the-art reinforcement learning algorithms which search in the space of actions [32]. Moreover, this approach results in static and linear policies, which can be efficiently deployed on real robots. The resulting policies are easy to store (they do not require storing large networks and other memory-intensive structures), and require only one matrix multiplication to compute the action at every time step. Furthermore, this algorithm works by searching in the space of model parameters, which makes it more suitable for long-horizon problems [18].

At every time step, we encode the state of the system $x_t$ as the cables lengths, current position, and current velocity of every node:

$$x_t = [l_t \quad r_t \quad \dot{r}_t] \qquad (3)$$

This representation is then multiplied by the matrix of the learned weights $\mathbf{W}(\theta)$ to produce the change of the target cables lengths $l_t^*$:

$$l_t^* = \mathbf{W}(\theta)x_t. \qquad (4)$$

Target cables lengths can then be sent to the lower level control system, such as pulley control system or to individual motor control systems.

## V. RESULTS AND TRAINING DYNAMICS

The simulation environment used in this paper is NASA Tensegrity Robotics Toolkit (NTRT) simulator, which uses Bullet Physics engine [3, 33]. For the description of the simulator, see papers [3, 33].

Our results are presented in Figure 2, where we report an average episode length in seconds (which proportional to the achieved reward) during the training phase. We also evaluate the found policy in different scenarios. First, we run the policy against a range of initial heights that are different from the fixed training height. Second, we assess the policy at lower decision frequencies. And third, we check whether the obtained controller can keep the structure to stand longer than the 20 seconds horizon, which was set at the training phase.

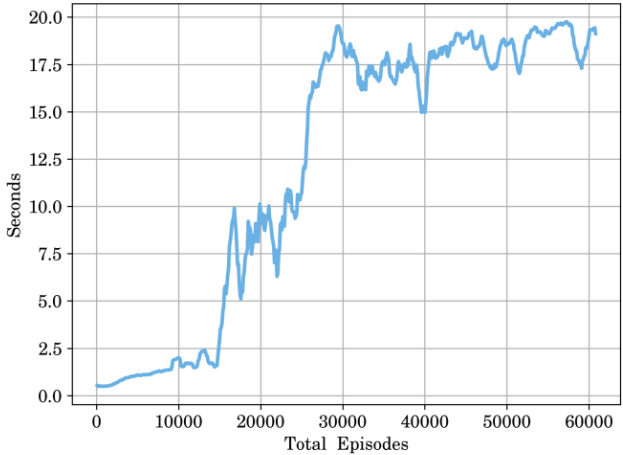

Fig. 2. Reward function (average time that the robot was able to remain stable), where 20 s is maximum possible reward.

Figure 2 shows average episode length over the training process, which represents the learning curve. The controller achieves the maximum possible reward around 30k episodes of training, thus successfully solving the task. We observe that after achieving the maximum reward, the policy does not degrade much and the average episode length stays around the same level. Figure 3 demonstrate the behavior of the best policy in the NTRT simulator.

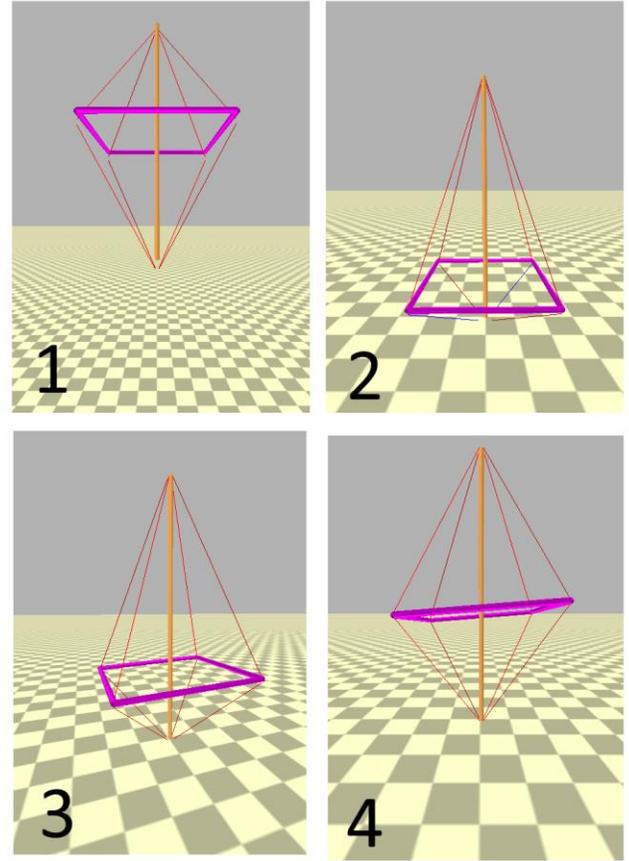

Fig. 3. Siulation frames during a single experiment with the hopper robot

Figure 3 shows the wide motion executed by the frame link of the robot (using the notation, introduced in Section III). Resulting stance of the robot (frame #4 in Figure 5) is stabilized.

One wanted property of the trained controller is an ability to keep the structure stand stable even for initial heights that were not sampled during the training phase. Figure 4 depicts the average episode length for different starting heights (experiments were repeated ten times for each height).

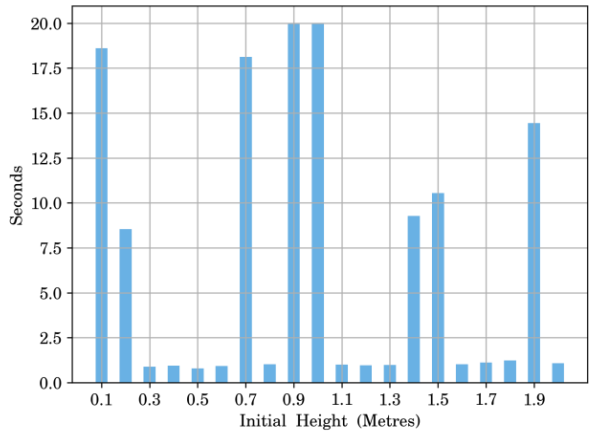

Fig. 4. Average episode length in evaluation, for different initial heights. The initial training height was fixed to 1 m.

We observe that a safe lower bound is only 0.1m, but if the structure is placed just above the ground it can succeed in most of the cases. We suggest that such results are possible when the observed states for different initial heights are close to the ones seen during the training. For instance, when falling from initial heights of around 0.1m, the robot can transfer directly to the state shown in frame #4 in Figure 3, which is has been seen by the robot during training.

The controller was trained on a relatively high control law update frequency: 1000H. In order to see how robust the found policy is to the change in frequency, we make experiments described above, while increasing the control action update interval (lowering the control update frequency). Figure 4 demonstrates the average episode length for a range of values using the controller trained on 1000Hz update frequency.

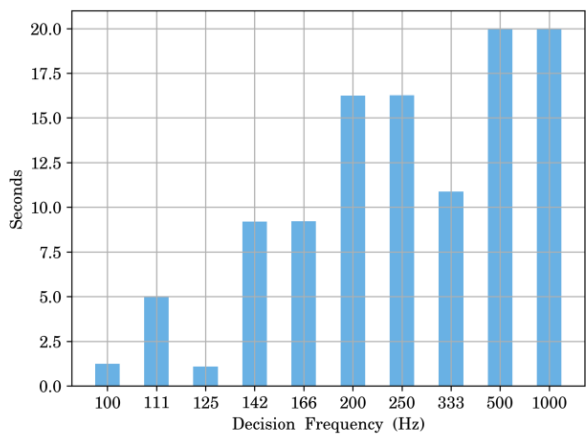

Fig. 5. Evaluation at lower decision frequencies. The controller was trained on the 1000Hz decision frequency.

We observe that lowering control frequency two fold (from 1000Hz to 500Hz) does not degrade the controller performance. Moreover, the controller succeeded in most of the cases even when the control update frequency was turned down to 200Hz. This result might be due to the fact that the change in observations (node velocities, accelerations, and cable lengths) at higher decision frequencies is relatively small. The static and linear nature of the policy only facilitates this phenomenon.

Additionally, we observe that the best policy can stand stable far longer than allowed during the training phase, up to ten minutes. During the successful episodes, the obtained policy converges to a specific set of states, where the structure is able to maintain vertical stability. Arriving at any state in this set allows for a successful evaluation on indefinitely long time ranges.

## VI. CONCLUSIONS

In this paper, we examined an application of Augmented Random Search to an open control problem - stabilization of a tensegrity hopper. Using this method, we could obtain static and linear policy that stabilizes the structure, achieving maximum reward at training phase, and keeps vertical stability far beyond the maximum training episode length. Moreover, the found policy is robust to the changes in control frequency. In particular, we showed that lowering the control frequency from 1000Hz to 500Hz does not lead to instability or performance degradation. We observed that the found policy demonstrates some robustness to different initial conditions (initial heights from which the robot falls). One of the directions on future research can be improving the robustness of the algorithm to a wider range of initial conditions, disturbances and parameter variations.